\documentclass{article} 
\usepackage{url}
\usepackage{graphicx}
\usepackage{amsmath,amssymb}
\newcommand{\bvec}[1]{\mbox{\boldmath $#1$}}
\title{Histogram Meets Topic Model: \\ Density Estimation by Mixture of Histograms}

\author{
Hideaki Kim\\
NTT Corporation, Japan\\
\texttt{kin.hideaki@lab.ntt.co.jp} \\
\and
Hiroshi Sawada\\
NTT Corporation, Japan\\
\texttt{sawada.hiroshi@lab.ntt.co.jp} \\
}

\begin{document}

\date{}

\maketitle

\begin{abstract}
The histogram method is a powerful non-parametric approach for estimating the probability density function of a continuous variable. But the construction of a histogram, compared to the parametric approaches, demands a large number of observations to capture the underlying density function. Thus it is not suitable for analyzing a sparse data set, a collection of units with a small size of data. In this paper, by employing the probabilistic topic model, we develop a novel Bayesian approach to alleviating the sparsity problem in the conventional histogram estimation. Our method estimates a unit's density function as a mixture of basis histograms, in which the number of bins for each basis, as well as their heights, is determined automatically. The estimation procedure is performed by using the fast and easy-to-implement collapsed Gibbs sampling. We apply the proposed method to synthetic data, showing that it performs well.
\end{abstract}

\section{Introduction}
Histogram, a non-parametric density estimator, has been used extensively for analyzing the probability density function of continuous variables \cite{rissanen1992, knuth2006, birge2006, chan2014}. Histograms are so flexible that they can model various properties of the underlying density like multi-modality, although they usually demand a large number of samples to obtain a good estimate. Thus the histogram method cannot be applied directly to a sparse data set, or a collection of units with a small data set. Due to the improvement of technology, however, it has recently become more important to analyze such diverse data of continuous variables as purchase timing \cite{kim2014,trinh2014}, period of word appearance \cite{wang2006}, check-in location \cite{cho2011} and neuronal spike time \cite{iyengar1997}, which could be sparse in many cases.

In this paper, by employing the probabilistic topic model called latent Dirichlet allocation \cite{blei2003}, we propose a novel Bayesian approach to estimating probability density functions. The proposed method estimates a unit's density function as a mixture of basis histograms, in which the unit's density function is characterized by a small number of mixture weights, alleviating the sparsity problem in the conventional histogram method. Furthermore, the model optimizes the bin width, as well as the heights, at the level of individual bases. Thus the model can implement a variable-width bin histogram as a mixture of regularly-binned histograms of different bin widths. We show that the estimation procedure in the proposed model can be performed by using the fast and easy-to-implement collapsed Gibbs sampling \cite{griffiths2004}. We apply the proposed method to synthetic data, clarifying that it performs well.

\section{Model}

\begin{table}
\centering
\caption{Notation \label{tb1}}
{\tabcolsep = 2.20mm
\begin{tabular}{ll} \hline
Symbol & Definition \rule[0mm]{0mm}{1mm}\\ \hline
$U$		& number of units \rule[0mm]{0mm}{1mm}\\
$N_u$ 	& number of variables in unit $u$ \rule[0mm]{0mm}{1mm}\\
$N$ 		& total number of variables \rule[0mm]{0mm}{1mm}\\
$K$ 		& number of basis histograms \rule[0mm]{0mm}{1mm}\\
$u$ 		& $u$th unit,~$1 \leq u \leq U$ \rule[0mm]{0mm}{1mm}\\
$t_j$ 	& $j$th variable in collection \rule[0mm]{0mm}{1mm}\\
$u_j$ 	& unit which generated $t_j$ \rule[0mm]{0mm}{3.25mm}\\
$z_j$ 	& latent variable of $t_j$,~$1\leq z_j \leq K$ \rule[0mm]{0mm}{1mm}\\
\hline
\end{tabular}
}
{\tabcolsep = 1.0mm
\begin{tabular}{ll} \hline
 &  \rule[0mm]{0mm}{1mm}\\ \hline
$T \equiv [T_0, T_1)$ 	& half-open range of variable \rule[0mm]{0mm}{1mm}\\
$x_l$ 	& lower boundary of $l$th bin \rule[0mm]{0mm}{1mm}\\
$W_k $ 	& number of bins in $k$th histogram \rule[0mm]{0mm}{1mm}\\
$\theta_{ku}$ 	& weight of $k$th histogram on unit $u$ \rule[0mm]{0mm}{1mm}\\
$\phi_{lk}$ 	& probability mass of $l$th bin \rule[0mm]{0mm}{1mm}\\
$ $ 			& in $k$th histogram,~$\sum_l \phi_{lk} = 1$ \rule[0mm]{0mm}{1mm}\\
$w_j^k$ 		& index of bin within which $t_j$ falls \rule[0mm]{0mm}{1mm}\\
$ $ 			& under $k$th bin number, $W_k$ \rule[0mm]{0mm}{1mm}\\
\hline
\end{tabular}
}
\end{table}

Suppose that we have a collection of $U$ units, each of which consists of $N_u$ continuous variables generated by each unit $u$ ($1\leq u \leq U$). For convenience, we number all the variables in the collection from $1$ to $N \equiv \sum_{u} N_u$ (in an arbitrary order), and define a set of collections, $\{ t_j \}_{j=1}^N$ and $\{ u_j \}_{j=1}^N$, where $t_j$ and $u_j$ are the $j$th variable and the unit which generated it, respectively. The notation is summarized in Table \ref{tb1}.
	
In our proposed model, it is assumed that each of the continuous variable, $t_j$, be generated from a mixture of histograms. In fact, as a generative process, a histogram can be described as a piecewise-constant probability density function \cite{knuth2006,endres2008}, which is a key point of our model construction. We first provide a description of the piecewise-constant distribution.

\subsection{Histogram: piecewise-constant distribution}
\begin{figure}
\begin{center}
\includegraphics{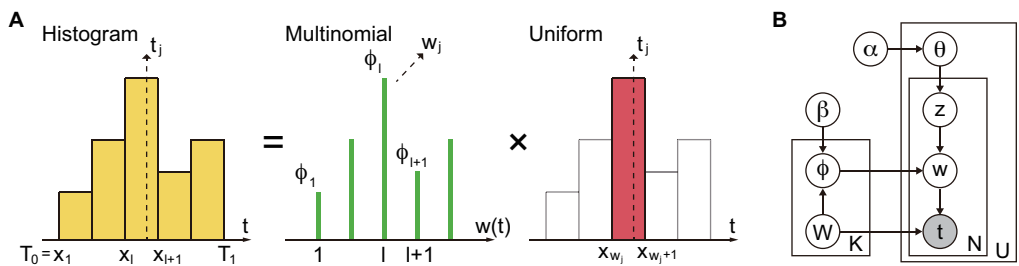} \\
\end{center}
\caption{(A) Histogram is described by a piecewise-constant probability density function. A sampling of $t_j$ from histogram can be implemented by the two steps: draw a bin index $w_j$ ($1\leq w_j \leq W$) from a multinomial distribution; and draw $t_j$ from the uniform distribution defined over the corresponding bin range $[x_{w_j}, x_{w_j+1})$. (B) Graphical model representation of HistLDA, a model which includes a histogram.}
\label{fig_Histlda}
\end{figure}
Histogram method describes an underlying density function by; (i) discretizing a half-open range of variable, $T=[T_0, T_1)$, into $W$ contiguous intervals (bins) of equal width, ($T_1 - T_0$)$/W$; and (ii) assigning a constant probability density, $h_l$, to each bin region, $[x_l, x_{l+1})$, for $1\leq l \leq W$. Here the lower boundary of bin, $x_l$, is given by, $x_l = T_0 + (l-1)(T_1 - T_0)/W$. In it, a continuous variable $t$ follows a piecewise-constant distribution,
\begin{equation}
p(t | h, W) = h_{w(t)}, \qquad \sum_{l=1}^W h_l~(T_1 - T_0) / W = 1,
\end{equation}
or equivalently, 
\begin{equation}\label{eq_hist}
p(t | \phi, W) = \phi_{w(t)}~W / (T_1 - T_0), \qquad \sum_{l=1}^W \phi_l = 1,
\end{equation}
where $\phi_l \equiv h_l~(T_1 - T_0) / W$ is the probability mass of the $l$th bin region $[x_l, x_{l+1})$, and $w(t)$ represents a discretization operator that transforms a continuous variable $t$ into the corresponding bin index, defined by
\begin{equation}\label{eq_hist2}
w(t) \equiv 1 + \text{int} \bigl[ W (t -T_0) / (T_1 - T_0) \bigr].
\end{equation}

It should be emphasized here that Eqs.~(\ref{eq_hist}-\ref{eq_hist2}) suggest that the observation process $p(t|\phi,W)$ can be decomposed into the following two processes (see Fig.~\ref{fig_Histlda}A): (i) Draw a bin index $w$ from a Multinomial distribution with parameter $\phi \equiv (\phi_1, \phi_2, \dots, \phi_W)$; (ii) Draw a continuous variable from an uniform distribution defined over the corresponding bin region, $[x_w, x_{w+1})$. Without the second process, a mixture model of $p(t|\phi,W)$ reduces to the original latent Dirichlet allocation (LDA) \cite{blei2003}, in which observed variables are discrete. In the next section, we construct a mixture of histograms by incorporating a uniform process into the original LDA. We call the proposed model the \textit{Histogram Latent Dirichlet Allocation} (HistLDA).

\subsection{Histogram latent Dirichlet allocation}
HistLDA estimates unit $u$'s probability density function as a mixture of basis histograms as follows:
\begin{equation}\label{eq_mix_hist}
p(t | \theta^{(u)}, \bvec{\phi}, \bvec{W}) = \sum_{k=1}^K \theta_{ku}~p(t | z = k, \phi^{(k)}, W_k),
\end{equation}
where $K$ is the number of mixture components, $z$ is a latent variable indicating the component from which the variable is drawn, $\bvec{W} \equiv \{W_k \}_{k=1}^K$ is the set of bin numbers, $\phi^{(k)} \equiv$ ($\phi_{1k}, \phi_{2k}, \dots, \phi_{W_k k}$) is the probability masses of the $k$th histogram, $\bvec{\phi} \equiv \{ \phi^{(k)} \}_{k=1}^K$ is its set, and $\theta^{(u)} \equiv$ ($\theta_{1u}, \theta_{2u}, \dots, \theta_{Ku}$) is the weights of the $K$ components on unit $u$. Each of the basis histograms, $p(t | z = k, \phi^{(k)}, W_k)$, is described by Eq.~(\ref{eq_hist}). Note that the set of basis histograms is shared by all the units, and heterogeneity across units is represented only through the weight $\theta^{(u)}$.

In accordance with LDA \cite{blei2003}, our HistLDA assumes the following generative process for a set of collections, $\{ t_j \}_{j=1}^N$ and $\{ u_j \}_{j=1}^N$:
\begin{enumerate}
\item For each basis histogram $k = 1, \dots, K$:
\begin{itemize}
\item[(a)] Draw number of bins $W_k \sim$ \text{Uniform} ($1, W_{\text{max}}$)
\item[(b)] Draw probability mass $\phi^{(k)} \sim$ \text{Dirichlet} ($\beta$, $W_k$)
\end{itemize}
\item For each unit $u = 1, \dots, U$:
\begin{itemize}
\item[(a)] Draw basis weight $\theta^{(u)} \sim$ \text{Dirichlet} ($\alpha$, $K$)
\end{itemize}
\item For each observation in collection $j = 1, \dots, N$:
\begin{itemize}
\item[(a)] Draw basis $z_j \sim$ Multinomial ($\theta^{(u_j)}$, $K$)
\item[(b)] Draw bin index $w_j \sim$ Multinomial ($\phi^{(z_j)}, W_{z_j}$)
\item[(c)] Draw variable $t_j \sim$ Uniform ($x_{w_j} , x_{w_j+1}$),
\end{itemize}
\end{enumerate}
where $\text{Uniform}(x,y)$ is the uniform distribution defined over an interval $[x,y]$, $\text{Dirichlet}(x,y)$ is the symmetric Dirichlet distribution of $y$ random variables with parameter $x$, $\text{Multinomial}(x,y)$ is the multinomial distribution of $y$ categories with equal choice probability $x$, $\alpha$ and $\beta$ are Dirichlet parameters, and $W_{\text{max}}$ is the maximum number of bins to be considered. 

The HistLDA is an extension of the original LDA in that (i) the number of bins (vocabulary), as a random variable, is not necessarily the same among the bases (topics), and (ii) observation is not a discrete bin index (word) drawn from a multinomial distribution, but a continuous variable further drawn from a uniform distribution defined over the corresponding bin's interval (see Fig.~\ref{fig_Histlda}). Also, it is worth noting that the HistLDA is a novel extension of Knuth's Bayesian binning model \cite{knuth2006} into Hierarchical structure.

Because being conjugate to Dirichlet priors, the multinomial parameters, $\theta^{(u)}$ and $\phi^{(k)}$, can be marginalized out from the generative model, leading to the joint distribution of data $\bvec{t} \equiv \{ t_j \}_{j=1}^N$, its latent basis $\bvec{z} \equiv \{z_j \}_{j=1}^N$, and the set of bin numbers $\bvec{W}$, as follows:
\begin{align}\label{eq_joint}
&p(\bvec{t}, \bvec{z}, \bvec{W} | \alpha, \beta) \notag \\
&= W_{\text{max}}^{-K} \prod_{u} \int d \theta^{(u)} p(\theta^{(u)} | \alpha) \prod_{j:u_j=u} p(z_j | \theta^{(u)}) \cdot \prod_{k} \int d \phi^{(k)} p(\phi^{(k)} | \beta) \prod_{j: z_j=k} p(t_j | z_j,\phi^{(k)},W_k) \notag \\
&= W_{\text{max}}^{-K} \prod_u \frac{\prod_k \Gamma (\alpha + N_{ku})}{\Gamma (K \alpha + N_u)} \frac{\Gamma (K \alpha )}{\Gamma (\alpha)^K} \prod_k \frac{\prod_{l=1}^{W_k} \Gamma (\beta + N_{kl})}{\Gamma (W_k \beta + N_k)} \frac{\Gamma (W_k \beta )}{\Gamma (\beta)^{W_k}} \biggl( \frac{W_k}{(T_1 - T_0)} \biggr)^{Nk},
\end{align}
where $\Gamma (x)$ is the gamma function, $N_{ku}$ is the number of times a variable of unit $u$ has been assigned to basis $k$, $N_{kl}$ is the number of times that a variable assigned to the $k$th basis histogram is addressed to the $l$th bin of the histogram, and $N_k = \sum_u N_{ku}$.

\subsection{Estimation by collapsed Gibbs sampling}
\begin{figure}
\begin{center}
\includegraphics{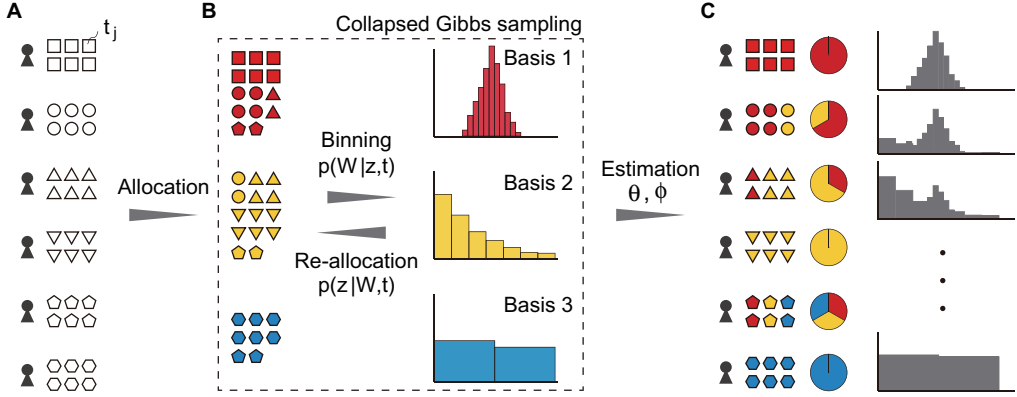} \\
\end{center}
\caption{A schema for the estimation procedure. (A) Each unit has a small size of continuous variables, which is indicated by each symbol (e.g. triangle). Here each of the data set is not enough to obtain a good estimate of each unit's density function. (B) Given the number of basis histograms (three in the figure) and their bin numbers $\bvec{W}$, according to the posterior, each of all the variables is allocated to one of the bases, colored in red, yellow and blue, respectively. Given the data set allocated to each basis, the bin number is optimized at the level of individual bases based on the posterior (Binning), and again the variables are re-allocated under the new bin numbers. The optimization is performed accurately due to the enough size of the data. The procedure is repeated before the joint posterior $(\bvec{z},\bvec{W}|\bvec{t})$ is converged. In collapsed Gibbs sampling, the heights of each histogram are not estimated explicitly, making the estimation easier to implement. (C) For a unit, each of the weights of the bases, $\theta^{(u)}$, are estimated by counting the proportion of the unit's data allocated to each basis, represented by a pie chart. Finally, the density function of each unit is estimated as a mixture of the basis histograms.}
\label{fig_schema}
\end{figure}

Given a collection of observed variables, $\bvec{t}$, we can estimate latent basis and number of bins, $\bvec{z}$ and $\bvec{W}$, based on the posterior distribution, $p(\bvec{z}, \bvec{W} | \bvec{t}, \alpha, \beta) \propto p(\bvec{t}, \bvec{z}, \bvec{W} | \alpha, \beta)$. In practice, by using the simple and easy-to-implement collapsed Gibbs sampling \cite{griffiths2004}, we obtain samples of $\bvec{z}$ and $\bvec{W}$ following $p($\bvec{z}$, $\bvec{W}$ |\bvec{t}, \alpha, \beta)$, from which the weight $\theta^{(u)}$ and the probability mass $\phi^{(k)}$, as well as the hyperparameters $\alpha$ and $\beta$, are estimated efficiently.    

\subsubsection*{Collapsed Gibbs sampling}
Given a set of bin numbers $\bvec{W}$ and the current state of all but one latent variable $z_j$, denoted by $\bvec{z}_{\backslash j}$, the basis assignment to $j$th variable is sampled from the following multinomial distribution:
\begin{equation}\label{eq_gibbs1}
p(z_j =k | \bvec{z}_{\backslash j}, \bvec{W}, \bvec{t}) \propto  (\alpha_k + N_{ku_j}^{\backslash j})~\frac{\beta + N_{kw^k_j}^{\backslash j}}{W_k \beta + N_{k}^{\backslash j}}~W_k, \qquad 1 \leq k \leq K,
\end{equation}
and given $\bvec{z}$ and all but one bin number $W_k$, denoted by $\bvec{W}_{\backslash k}$, the $k$th bin number $W_k$ is sampled from the following multinomial distribution:
\begin{equation}\label{eq_gibbs2}
p(W_k | \bvec{z}, \bvec{W}_{\backslash k}, \bvec{t}) \propto \frac{\prod_{l=1}^{W_k} \Gamma (\beta + N_{kl})}{\Gamma (W_k \beta + N_k)} \frac{\Gamma (W_k \beta )}{\Gamma (\beta)^{W_k}} W_k^{N_k}, \qquad 1 \leq W_k \leq W_{\text{max}},
\end{equation}
where $N_{\cdot}^{\backslash j}$ represents the count that does not include the current assignment of $z_j$. Here $w^{k}_j$ represents the $k$th histogram's bin index within which $t_j$ falls, defined as
\begin{equation}  
w^{k}_j \equiv 1 + \text{int} \bigl[ W_k (t_j -T_0) / (T_1 - T_0) \bigr].
\end{equation}
Eqs. (\ref{eq_gibbs1}) and (\ref{eq_gibbs2}) are easily derived from Eq. (\ref{eq_joint}) (not  shown here).

In each sampling of $\bvec{z}$ and $\bvec{W}$, the hyperparameters of the Dirichlet priors, $\alpha$ and $\beta$, can be updated by using the fixed-point iteration method described in \cite{minka2000} as follows:
\begin{equation}\label{eq_hyper_it}
\begin{split}
\alpha &\leftarrow \alpha~\frac{\sum_u \sum_k \Psi (\alpha + N_{ku})  - U K \Psi(\alpha)}{K \sum_u \Psi (K \alpha+N_u)- U K \Psi(K \alpha) },\\
\beta &\leftarrow \beta~\frac{ \sum_k \sum_l \Psi(\beta + N_{kl})  - \bigl( \sum_k W_k \bigr) \Psi(\beta)}{\sum_k W_k \Psi (W_k \beta + N_k) - \sum_k W_k \Psi(W_k \beta) },
\end{split}
\end{equation}
where $\Psi (x)$ is the digamma function defined by the derivative of $\log \Gamma (x)$. For details, see Appendix \ref{ap1}.

In practice, we set the initial hyperparameters as $\alpha^{(0)}$ = $\beta^{(0)}$ = 0.5 (Jeffreys non-informative prior \cite{jeffreys1961}), and draw the initial $z_j \in \bvec{z}$ from $\text{Dirichlet} (\alpha^{(0)},K)$. The algorithm of the collapsed Gibbs sampling in HistLDA is summarized in Algorithm 1 (see also Fig.~\ref{fig_schema}B).

\subsubsection*{Estimation of basis histograms and their weights on units}
By repeating the collapsed Gibbs sampling (\ref{eq_gibbs1}-\ref{eq_hyper_it}) before the convergence is achieved, we first estimate the hyperparameters, $\hat{\alpha}$ and $\hat{\beta}$, as the last updated values. At the same time, we also estimate the set of bin numbers, $\hat{\bvec{W}} \equiv \{ W_k \}_{k=1}^K$, as the last updated values. Next, given $\hat{\alpha}$, $\hat{\beta}$ and $\hat{\bvec{W}}$, we further draw $N_p$ samples of basis assignment, [$\bvec{z}^{(1)}, \bvec{z}^{(2)}, \dots, \bvec{z}^{(N_p)}$], according to Eq. (\ref{eq_gibbs1}), and obtain the posterior mean estimate of the weight $\theta$ and probability mass $\phi$ as follows:
\begin{equation}\label{eq_param}
\hat{\theta}_{ku} \simeq \frac{1}{N_p} \sum_{p=1}^{N_p} \frac{\hat{\alpha} + N_{ku}^{(p)}}{K \hat{\alpha} + N_u},\qquad
\hat{\phi}_{l k} \simeq \frac{1}{N_p} \sum_{p=1}^{N_p}  \frac{\hat{\beta} + N_{k l}^{(p)}}{\hat{W_k} \hat{\beta} + N_k^{(p)}},
\end{equation}
where $N_{ku}^{(p)}$, $N_{kl}^{(p)}$ and $N_{k}^{(p)}$ represent the sufficient statistics in the $p$th sample, $\bvec{z}^{(p)} \equiv \{ z_j^{(p)} \}_{j=1}^N$. In the following experiment, $N_p$ was set at $100$.

In Figure \ref{fig_schema}, we give an intuitive explanation of the estimation procedure in HistLDA.

\begin{table}
\centering
\label{tb2}
\begin{tabular}{l} \hline
{\bf Algorithm 1} Collapsed Gibbs sampling in HistLDA \\ \hline 
Set $K$ and $T$, and initialize $\alpha = \beta = 0.5$, and $W_k = 1$ for $1 \leq k \leq K$. \\
Draw $\bvec{z}$ from Dirichlet ($\alpha$, $K$). \\
Count sufficient statistics $N_{k l}$, $N_{k u}$ and $N_{k}$ for $1 \leq l \leq W_k$, $1 \leq k \leq K$.\\
{\bf repeat} \\
\quad {\bf for} $k=1$ to $K$ {\bf do}\\
\qquad Draw $W_k$ from Eq. (\ref{eq_gibbs2}). \\
\qquad Update $N_{k l}$ for $1 \leq l \leq W_k$, under a new $W_k$. \\
\quad {\bf for end} \\
\quad {\bf for} $j=1$ to $N$ {\bf do}\\
\qquad Set $N_{z_j u} = N_{z_j u} -1$, ~$N_{z_j w_j^{z_j}} = N_{z_j w_j^{z_j}} -1$, ~$N_{z_j} = N_{z_j} - 1$. \\
\qquad Draw $z_j$ from Eq. (\ref{eq_gibbs1}). \\
\qquad Set $N_{z_j u} = N_{z_j u} +1$, ~$N_{z_j w_j^{z_j}} = N_{z_j w_j^{z_j}} +1$, ~$N_{z_j} = N_{z_j} + 1$. \\
\quad {\bf for end} \\
\quad Update $\alpha$ and $\beta$ based on Eq. (\ref{eq_hyper_it}). \\
{\bf until} posterior $p(\bvec{z}, \bvec{W} | \bvec{t}, \alpha, \beta)$ is converged. \\
\hline
\end{tabular}
\end{table}

\section{Result}
\begin{figure}
\begin{center}
\includegraphics{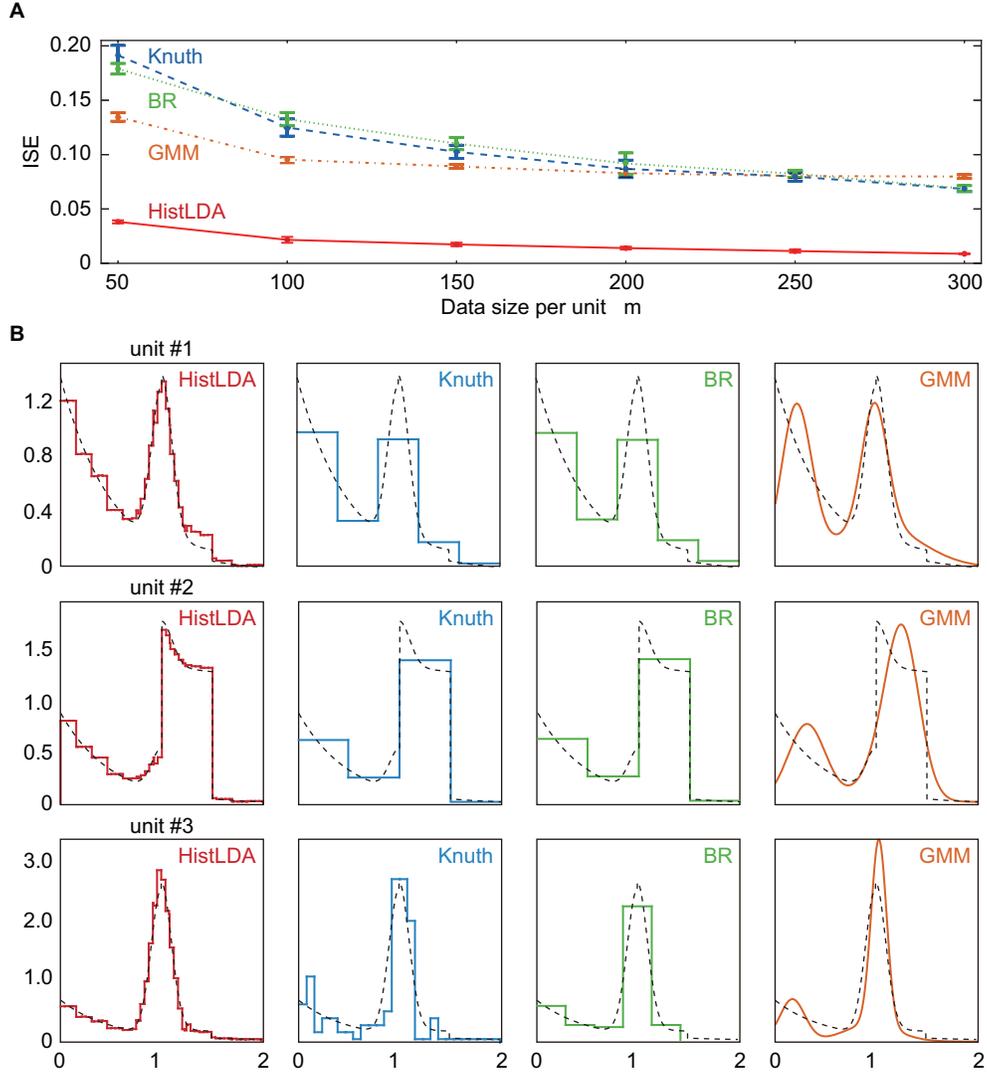} \\
\end{center}
\caption{Density estimation with synthetic data. (A) The ISE between the intended and the estimated density functions against the data size per unit $m$. The error bars represent standard deviations of ISE when the density estimation was performed three times using the three sets of data generated. (B) Three units' examples of estimated probability density functions for $m=100$. The solid line represents the density function estimated by each method, and the dashed line represents the true density function. }
\label{fig_ISE}
\end{figure}

To confirm that the HistLDA works well on a sparse data set, that is, a collection of units consisting of a small size of variables, we evaluated the performance of HistLDA together with the reference methods on synthetic data.  

As the reference methods, we adopted two histogram methods, the Bayesian binning method proposed by Knuth \cite{knuth2006} (Knuth) and penalized-maximum likelihood method by Birg\'{e} and Rozenholc \cite{birge2006} (BR), and a parametric method, Gaussian mixture model (GMM). The three methods estimate a unit's probability density function based on its own observed variables.  

We made synthetic data in the scenario that a unit generated continuous variables from a complex probability density function comprising the following three different types of distributions: the normal distribution with mean $1$ and variance $0.1^2$, the exponential distribution with rate parameter $2$, and the uniform distribution defined over the interval $[1, 1.5]$. Here each unit was characterized by the mixing proportions, which were sampled from a uniform or flat Dirichlet distribution with respect to each unit.  Generating a collection comprising $U = 100$ units, each of which had $m$ variables, we estimated the units' underlying probability density functions from the data, and evaluated the goodness of the estimation in terms of the average integrated squared error (ISE) of probability density function:
\begin{equation}
\text{ISE} = \frac{1}{U} \sum_{u=1}^U \int_{T_0}^{T_1} \bigl( \hat{p}_u(t) - p_u(t) \bigr)^2 dt,
\end{equation}
where the range of variable, $T\equiv [T_0, T_1)$, was set at $[0,2)$, and $p_u (t)$ and $\hat{p}_u(t)$ represent the intended and the estimated probability density functions, respectively. In the experiment, the number of mixture components was set at three for both HistLDA and GMM. The data size per unit, $m$, was specified as $m = 50$, $100$, $150$, $200$, $250$, $300$. 

Figure \ref{fig_ISE}A compares HistLDA's ISE against the results achieved by the reference methods, demonstrating that HistLDA performed better than the other methods for all the data size per unit, $m$. The comparison between HistLDA and the other histogram methods (Knuth and BR) found that HistLDA performed relatively much better even in the small $m$. This suggests that our HistLDA copes with the sparsity problem in the usual histogram methods. Also, Figure \ref{fig_ISE}A shows that the histogram methods (HistLDA, Knuth and BR) performed better when the data size was larger, while the performance of GMM was not improved significantly. Parametric approaches like GMM, which work robustly in sparse data, usually perform poorly under the wrong assumption of underlying distribution. In the experiment, the underlying density function of each unit was far from Gaussian (see Fig.~\ref{fig_ISE}B). 

\begin{figure}
\begin{center}
\includegraphics{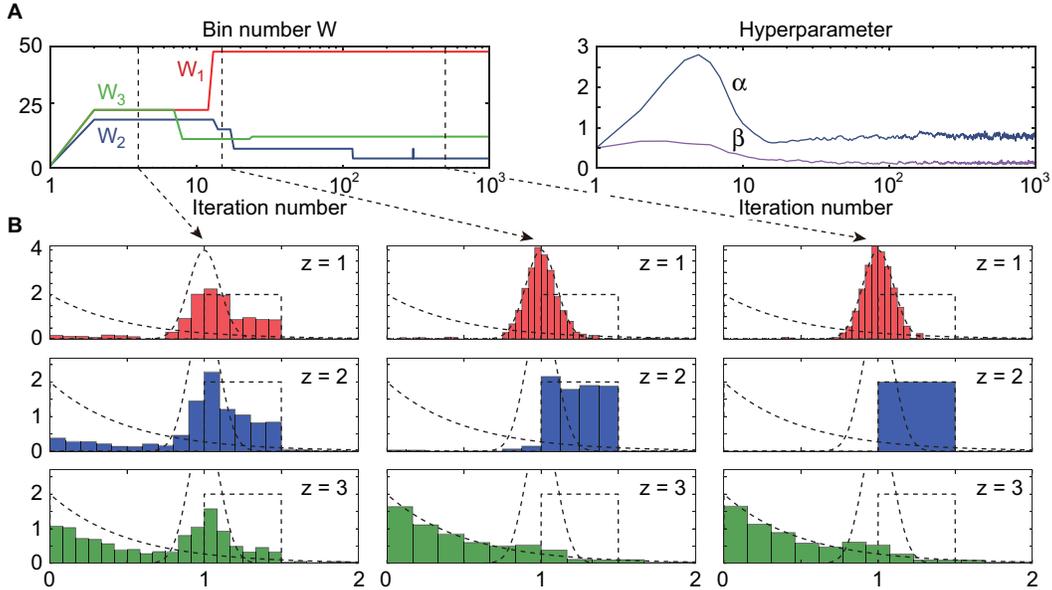} \\
\end{center}
\caption{Collapsed Gibbs sampling with synthetic data. (A)  Plots of bin number $\bvec{W} = ( W_1, W_2, W_3 )$ and hyperparameter ($\alpha$, $\beta$) versus iteration number. (B) The sampled bin number and the estimated basis histograms at the 4th, 10th, and 500th iteration. In each figure, the three distributions depicted by the dashed lines represent the underlying normal, exponential, and uniform distributions.}
\label{fig_estimation}
\end{figure}

Figure~\ref{fig_ISE}B shows three units' examples of estimated probability density functions for $m$ = $100$. Each unit had a complex and unit-specific distribution, but HistLDA obtained a good estimate of each distribution by adopting small bin widths (large number of bins). Generally, the bin width of histogram is estimated to be smaller in a larger data set \cite{shimazaki2007}, of which situation is realized in HistLDA by allocating the whole data of all the units into a small number of basis histograms (see Fig.~\ref{fig_schema}). Furthermore in Fig.~\ref{fig_ISE}B, HistLDA seems to have adjusted bin widths depending on the location: large bin width was used around 0, and small one being used around 1. HistLDA implements a variable-width bin histogram by way of a mixture of regular histograms with different bin widths. Figure~\ref{fig_estimation} displays how the bin number (bin width) for each basis histogram was optimized in the collapsed Gibbs sampling.

\section{Summary}
We have proposed a novel histogram density estimation that overcomes the sparsity of data, by incorporating the latent Dirichlet allocation  (LDA) into the histogram method. The proposed method estimates a density function as a mixture of histograms, of which bin numbers or bin widths are optimized together with their heights, at the level of individual histograms. By way of a mixture of regularly-binned histograms of different bin widths, it can implement a variable-width bin histogram. As with the LDA, all the estimation procedure is performed by using the fast and easy-to-implement collapsed Gibbs sampling. We assessed the goodness of the proposed method by examining synthetic data, and demonstrated that it performed well when data sets were sparse.

\appendix
\section{Estimation of hyperparameters}\label{ap1}
Based on the empirical Bayes method, the hyperparameters in the Dirichlet distributions, $\alpha$ and $\beta$, can be estimated by maximizing the marginal likelihood,
\begin{equation}
p(\bvec{t} | \alpha, \beta) = {\textstyle \sum_{\bvec{z}, \bvec{W}}}~p(\bvec{t}, \bvec{z}, \bvec{W}| \alpha, \beta).
\end{equation}
The maximization is performed by using the Monte Carlo EM algorithm, leading to the following update rule, 
\begin{equation}
\alpha^{+}, \beta^{+} = \arg \max_{\alpha, \beta} E_{\bvec{z},\bvec{W}} \bigl[ \log p(\bvec{t}, \bvec{z}, \bvec{W} | \alpha, \beta)~|~\bvec{t},\alpha^{-}, \beta^{-} \bigr],
\end{equation}
where $\alpha^{+}$ and $\beta^{+}$ are the updated values of hyperparameters, and $E_{\bvec{z},\bvec{W}} \bigl[ \cdot |~\bvec{t},\alpha^{-}, \beta^{-} \bigr]$ represents the expectation with respect to the posterior distribution of $\bvec{z}$ and $\bvec{W}$ under the previous estimate of the hyperparameters, ($\alpha^{-}, \beta^{-}$). This time, we employ the stochastic EM, a variant of Monte Carlo EM algorithm, in which the expectation is replaced by a sample taken from the posterior distribution (\ref{eq_gibbs1}-\ref{eq_gibbs2}), as
\begin{equation}\label{eq_maxmax}
\alpha^{+}, \beta^{+} = \arg \max_{\alpha, \beta} \log p(\bvec{t}, \bvec{z}^{-}, \bvec{W}^{-}| \alpha, \beta),
\end{equation}
where $\bvec{z}^{-}$ and $\bvec{W}^{-}$ are the values of $\bvec{z}$ and $\bvec{W}$ sampled by the collapsed Gibbs sampling with the previous hyperparameters, ($\alpha^{-}, \beta^{-}$). Although having no analytical forms, Eq.~(\ref{eq_maxmax}) can be computed by iterating the following fixed-point equation \cite{minka2000}:  
\begin{align}
\alpha^{+} &\leftarrow \alpha^{+} \frac{\sum_u \sum_k \Psi(\alpha^{+}+N_{ku}^{-})  - U K \Psi(\alpha^{+})}{K \sum_u \Psi (K \alpha^{+}+N_u)- U K \Psi(K \alpha^{+}) },\\
\beta^{+} &\leftarrow \beta^{+} \frac{ \sum_k \sum_l \Psi(\beta^{+} + N^{-}_{kl})  - \bigl( \sum_k W_k^{-} \bigr) \Psi(\beta^{+})}{\sum_k W_k^{-} \Psi (W_k^{-} \beta^{+} + N^{-}_k) - \sum_k W_k^{-} \Psi(W_k^{-} \beta^{+}) },
\end{align}
where $\Psi (x)$ is the digamma function defined by the derivative of $\log \Gamma (x)$, and $N_{ku}^{-}$, $N_{kl}^{-}$ and $N_{k}^{-}$ represent the realization of $N_{ku}$, $N_{kl}$ and $N_{k}$ in the collapsed Gibbs sampling with the previous hyperparameters, ($\alpha^{-}, \beta^{-}$), respectively.

\small{

}

\end{document}